\pdfoutput=1
\documentclass[11pt]{article}

\usepackage[]{acl}

\usepackage{times}
\usepackage{latexsym}

\usepackage[T1]{fontenc}
\usepackage[utf8]{inputenc}
\usepackage{amsmath,amssymb}
\usepackage{microtype}
\usepackage{enumitem}
\usepackage{inconsolata}
\usepackage{todonotes}
\usepackage[normalem]{ulem}
\usepackage{booktabs}
\usepackage{subcaption}
\usepackage{graphicx}
\usepackage{multirow}
\usepackage{comment}
\usepackage[frozencache,cachedir=.]{minted}

\usepackage{listings}
\usepackage{color}

\usepackage{pifont}% http://ctan.org/pkg/pifont

\usepackage{enumitem}
\setlist{nosep,topsep=-\parskip}

\definecolor{blue}{rgb}{0,0, 0.6}
\definecolor{dkgreen}{rgb}{0,0.6,0}
\definecolor{gray}{rgb}{0.5,0.5,0.5}
\definecolor{mauve}{rgb}{0.58,0,0.82}
\definecolor{mauve}{rgb}{0,0,0}
\definecolor{black}{rgb}{0,0,0}

\usepackage{inconsolata}

\usepackage{soul}

\definecolor{lightblue}{rgb}{.50,.90,0.51}
\definecolor{tri}{rgb}{.25,.88,.82}
\definecolor{lilac}{rgb}{0.85,0.64,0.85}
\definecolor{atomictangerine}{rgb}{1.0, 0.6, 0.4}

\newcommand{\benchpack}{LLMeBench}

\newcommand{\code}[1]{\texttt{#1}}

\title{\benchpack: A Flexible Framework for Accelerating LLMs Benchmarking}

\author{
Fahim Dalvi, Maram Hasanain,  Sabri Boughorbel, Basel Mousi, Samir Abdaljalil, \\
 \textbf{Nizi Nazar, Ahmed Abdelali\thanks{~~The contribution was made while the author was at the Qatar Computing Research Institute.}, Shammur Absar Chowdhury,} \\
\textbf{Hamdy Mubarak, Ahmed Ali, Majd Hawasly, Nadir Durrani, Firoj Alam} \\
Qatar Computing Research Institute, HBKU, Qatar\\
\{faimaduddin,fialam\}@hbku.edu.qa
}

\begin{document}
\maketitle
\begin{abstract}
The recent development and success of Large Language Models (LLMs) necessitate an evaluation of their performance across diverse NLP tasks in different languages. Although several frameworks have been developed and made publicly available, their customization capabilities for specific tasks and datasets are often complex for different users. In this study, we introduce the \benchpack{}\footnote{\textbf{LLM} \textbf{e}ffectiveness \textbf{Bench}marking. Can be pronounced as ``lemme bench''.} framework, which can be seamlessly customized to evaluate LLMs for any NLP task, \textit{regardless of language}. 
The framework features generic dataset loaders, several model providers, and pre-implements most standard evaluation metrics. It supports in-context learning with zero- and few-shot settings. A specific dataset and task can be evaluated for a given LLM in less than 20 lines of code while allowing full flexibility to extend the framework for custom datasets, models, or tasks. The framework has been tested on 31 unique NLP tasks using 53 publicly available datasets within 90 experimental setups, involving approximately 296K data points. We open-sourced \benchpack{} for the community\footnote{\url{https://github.com/qcri/LLMeBench/}} and a video demonstrating the framework is available online.\footnote{\url{https://youtu.be/9cC2m_abk3A}}
\end{abstract}

% Initially developed to evaluate Arabic NLP tasks using OpenAI's GPT and BLOOM models; it can be seamlessly customized for any NLP task and model, \textit{regardless of language}. 

\section{Introduction}
\label{sec:introduciton}

% The rapid advancement of sophisticated large language models (LLMs), supported by in-context learning (\cite[ICL]{dong2023survey}), has gained unprecedented popularity among both the research and development community. The emergence of such large models with ICL paradigm facilitated diverse applications such as solving mathematical reasoning problems \cite{wei2023chainofthought}.
% % without a need for exhaustive computational resources. 
% However, systematic evaluation and comparison against state-of-the-art is important to accurately gauge the potential of these LLMs.  A comprehensive evaluation allows understanding of the strengths and weaknesses of these models; guides us towards better human-LLMs interactions through prompting; and facilitates their broader applicability in different scenarios, especially in domains where safety and security are paramount concerns (e.g., healthcare, financial institutes) \cite{chang2023survey, zhao2023survey, zhu2023promptbench}.

\begin{figure}[t]
\centering
\includegraphics[width=\columnwidth]{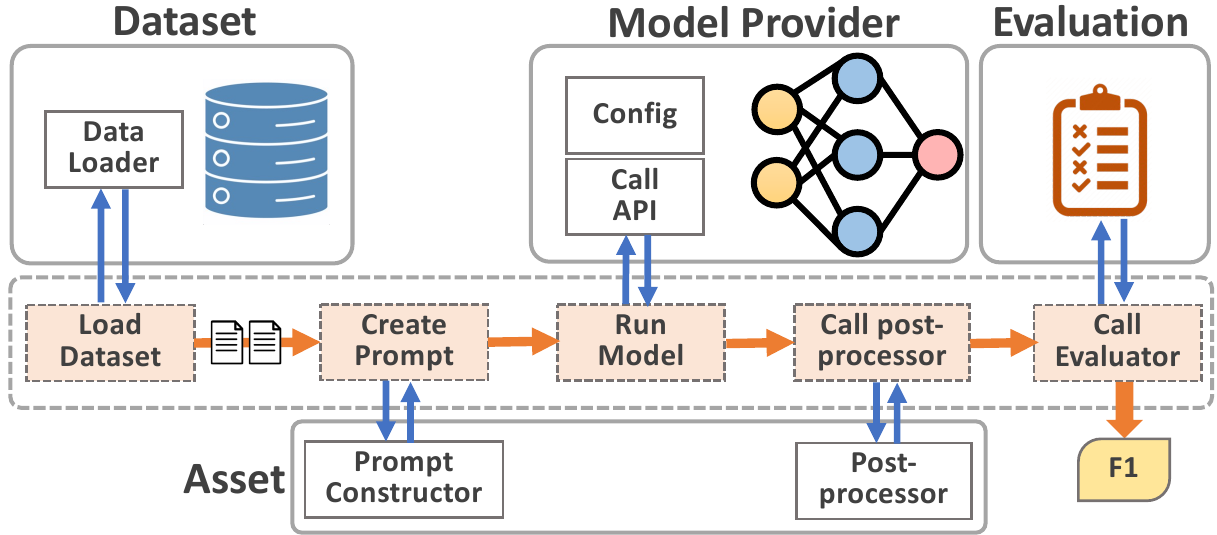}
\vspace{-5pt}
\caption{The architecture of the \benchpack{} framework.
% Overview of the \benchpack{} architecture. %Modules with dotted boarders are the core of the framework, which is already implemented. 
The dotted boxes represent the core implemented modules of the architecture. %, which is already implemented. 
Customization for new tasks, datasets, and models can be done on \code{Dataset}, \code{Model Provider}, \code{Evaluation}, and \code{Asset} modules.
% Modules with dotted boarders are already included in the framework architecture.
% \firoj{@Maram, all modules are already included, no?}\maram{Yes, I was trying to differentiate between modules implemented in the framework (core) and those the user has to implement.}
}
\label{fig:pipeline}
\vspace{-15pt}
\end{figure}

The rapid advancement of sophisticated large language models (LLMs), supported by in-context learning (ICL) \cite{dong2023survey}, has gained unprecedented popularity among both the research and development communities. The emergence of such large models %with the ICL paradigm 
facilitated diverse applications such as solving mathematical reasoning problems~\cite{weichain}. Given their success, a systematic evaluation and comparison against state-of-the-art is important to accurately gauge the potential of LLMs. A comprehensive evaluation allows understanding of the strengths and weaknesses of these models; guides us towards better human-LLMs interactions through prompting; and facilitates their broader applicability in different scenarios, especially in domains where safety and security are paramount concerns (e.g., healthcare, financial institutes)~\cite{chang2023survey, zhao2023survey, zhu2023promptbench}.

% To explore the LLMs capability and generalization potential for different domains, tasks, and languages, we are in urgent need of a unified evaluation benchmarking framework. This need arises because evaluation assists us in understanding their strengths and weaknesses, guides us towards better human-LLMs interactions through prompting, and facilitates their broader applicability in different scenarios, especially in domains where safety and security are paramount concerns (e.g., healthcare, financial institutes) \cite{chang2023survey, zhao2023survey, zhu2023promptbench}.

Numerous initiatives were launched to comprehensively assess the performance of LLMs on standard NLP tasks. The HELM project \citep{liang2022holistic} conducted a thorough evaluation of LLMs for English, spanning various metrics and scenarios. Additionally, the BIG-Bench initiative \citep{srivastava2023beyond} introduced an extensive evaluation of 214 tasks, even encompassing languages with limited resources. Notably, evaluations have been carried out on models like GPT2.5~\cite{radford2019language}, ChatGPT~\cite{openai2023gpt4}, and BLOOM~\cite{scao2022bloom} within multitask, multilingual, and multimodal settings. These evaluations were further extended to low-resource languages~\citep{bang2023multitask, ahuja2023mega, hendy2023good, khondaker2023gptaraeval}.

% Several initiatives are taken to evaluate the performance of the LLMs on standard NLP tasks. The HELM project \citep{liang2022holistic} assessed English LLMs across various metrics and scenarios. BIG-Bench \citep{srivastava2022beyond} introduced a large-scale evaluation with 214 tasks, including low-resource languages. GPT2.5~\cite{radford2019language}, ChatGPT~\cite{openai2023gpt4}, and BLOOM~\cite{scao2022bloom}, are also evaluated for multitask, multilingual, multimodal settings including the evaluation in low-resource language settings \citet{bang2023multitask,ahuja2023mega,hendy2023good,khondaker2023gptaraeval}. 
% hendy2023good
% \textcolor{red}{XYZ developed a closed evaluation framework with 15 distinct tasks from cross-lingual benchmarking. Other efforts include X, Y which depends on dataset hosted on the hugging face platform.  WILL EXTEND Later}
% To make the benchmarking effort accessible, 

% Seamlessly evaluating such LLMs for many diverse tasks often poses challenges in terms of cost, effort and time due to the complexities such as managing API calls, adding tasks, datasets, evaluation measures and/or hosting datasets in the public platform (e.g., Hugging Face). To overcome such limitations, in this study, we introduce ``\benchpack{}'': \textbf{LLM} \textbf{e}ffectiveness \textbf{Bench}marking, to promote comprehensive evaluation of these LLMs with seamless and flexible implementation. The proposed framework, shown in Figure \ref{fig:pipeline}, allows the users to test any available LLMs and integrate customized tasks, datasets and evaluation metrics. 

Evaluating LLMs across diverse tasks often entails challenges related to costs, effort, and time due to complexities like handling API calls, task integration, dataset inclusion, evaluation measures, and potentially hosting datasets on public platforms (e.g., Hugging Face (HF)). To overcome these limitations, in this study, we introduce ``\benchpack{}'', which facilitates a comprehensive evaluation of these LLMs through a seamless and flexible implementation. The proposed framework, as depicted in Figure \ref{fig:pipeline}, empowers users to assess various LLMs while simplifying the integration of custom tasks, datasets, and evaluation metrics.

% while maintaining data privacy in terms of a need to host any public platform (e.g., Hugging Face).
% \textcolor{red}{in terms of a need to host any public platform (e.g., Hugging Face).} 
% \maram{Like the idea of data privacy, but not sure the prev. sentence is clear} 
% \textbf{L}arge \textbf{L}anguage \textbf{M}odel-Based \textbf{E}ffective \textbf{Bench}marking, 
% The modular framework supports diverse task-specific pre-/post-processing, and evaluation metrics in both zero- and few-shot settings and can be further customized as per need. 

% LLMeBench distinguishes itself from its contemporaries through the following features:

A few evaluation frameworks have emerged to facilitate extensive benchmarking of LLMs. Among these are OpenAI evals,\footnote{\url{https://github.com/openai/evals}} LM Harness~\cite{eval-harness}, and OpenICL~\cite{wu-etal-2023-openicl}. Each framework offers functionalities tailored to specific requirements. For instance, OpenICL focuses on few-shot learning techniques. Our contribution, \benchpack{}, stands out by emphasizing a user-friendly, plug-and-play design, that can seamlessly integrate into existing experimental workflows, setting it apart from other alternatives. \benchpack{}'s uniqueness lies in the following features:

\begin{itemize} %[leftmargin=*, nosep, topsep=0pt, after=\vspace{0pt}]
    \item Supports several generic data loaders (e.g., HF datasets), pre-implements several model providers (such as OpenAI and HF inference APIs for remote execution, and FastChat~\cite{zheng2023judging} and Petals~\cite{borzunov2022petals} for local deployments), and supports all standard tasks and evaluations, such as classification, regression, etc. Evaluating a new task/dataset/model can be done in as few as 20 lines.
    \item Allows the user to create their own data loader, connecting to their local server, ensuring data privacy and security.
    \item Provides users with the flexibility to design diverse tasks, allowing customization of data input/output formats and evaluation criteria.
    \item Supports zero- and few-shot learning paradigms with $\sim$300 zero-/few-shot prompts serving as a valuable community resource.
    \item Enables automatic selection of few-shot examples from a user-defined train/dev set using a maximal marginal relevance-based approach. %\cite{carbonell1998use}.
    \item Implements an efficient caching mechanism to prevent repeated API calls, resulting in cost savings and resolution of time-out issues. 
    \item Offers extensive logging and caching capabilities, allowing iterative model outputs post-processing.
    \item Provides an auto-download mechanism for public datasets, accelerating experimentation.
    \item Includes 31 tasks recipes featuring different model providers. Rigorously tested with 53 datasets associated with 12 languages.
    % \item Provides 140 zero-/few-shot prompts, serving as a valuable community resource. 
\end{itemize}

\noindent %Moreover, 
Furthermore, \benchpack{} is an open-source, user-friendly, and adaptable comprehensive benchmarking framework for LLMs. It empowers both experts and non-experts to assess conventional and unique NLP tasks, enhancing comprehension of the models' capabilities and their applicability across standard and novel tasks.

% What problem does the proposed system address?
% Why is the system important and what is its impact?
% What is the novelty in the approach/technology on which this system is based?
% Who is the target audience?
% How does the system work?
% How does it compare with existing systems?
% How is the system licensed?

%\section{Framework}
\section{\benchpack}
\label{sec:framework}

% Figure~\ref{fig:pipeline} provides an overview of the \benchpack{} architecture. This framework was designed with a focus on modularity, extensibility, module reusability, and task independence. To spare users from having to implement common elements across experimental setups but not specific to a task, the architecture was designed to expect a unified format of input and intermediate outputs. This was achieved by implementing a pipeline that uses key-value dictionaries to pass data through.
% To alleviate
% The framework was designed with modularity in mind to allow for extensibility, reusability of modules, and independence across tasks. 
%The architecture was also designed such that the user is spared the burden of implementing elements common across experimental setups but not specific to a task. 
% This was achieved by expecting a unified format of input and intermediate outputs through the use of key-value dictionaries passing through the pipeline. 

In Figure~\ref{fig:pipeline}, we provide the architecture of the \benchpack{} framework. To ease the burden on users in implementing common elements across experimental setups, which are not specific to a task, the architecture was designed to support a uniform format for both input and intermediate outputs. This was achieved by employing a pipeline that utilizes key-value dictionaries to seamlessly pass data. The framework incorporates \textit{four} fundamental modules, discussed below. %elaborated upon below. Overseeing their invocation and intercommunication is the \code{Driver}. 
%It also includes a \code{Driver} that controls invocation and communication among them.
The process starts with a \code{Dataset}, where each input sample $S_i$ is %passed 
routed to the \code{Asset} module. Within this module, a prompt is created 
% generated 
and then passed to the \code{Model Provider} for processing. The model's response is then funneled back to the \code{Asset} module for post-processing. 
%The response from the model is subsequently returned to the \code{Asset} module for post-processing. 
As the processing of all input samples and the generation of corresponding responses conclude, the \code{Evaluation} module takes on the task of computing evaluation metrics.
The whole process of intercommunication is overseen by a \code{Benchmark Driver}.
%Once all input samples have been processed and responses generated, the \code{Evaluation} module carries out the evaluation. 
Throughout these processes, the inputs, processed data, and the intermediate outputs are cached for re-use and quick experimentation. 
% Further details on each of these components are provided in the following sub-sections.
% For a comprehensive understanding of each component, plese the following sub
% let's delve into the forthcoming sub-sections.
%All these processes are supported by a cache system that stores the outputs of the processed inputs. Further details on each of these components is provided in the following sub-sections.

% Given the core four modules described next, the framework offers a \code{Driver} that controls invocation and communication across them. 
% Starting from a \code{Dataset}, the pipeline passes each input sample $S_i$ to the \code{Asset} module, through which the prompt is created. 
% The prompt is then passed to the \code{Model} to run it, then the model response is passed to the \code{Asset} to be post-processed. 
% Finally, given responses for all input samples, evaluation is done through the \code{Evaluation} module. All this functionality is backed up by a cache to store the output of successfully processed inputs. 

% \todo[inline]{Do we need to talk about design principles? like modularity, efficiency, generality etc... }
% describing overall pipeline, flow of data, architecture figure, and a walking example

\begin{listing}[!t]
\footnotesize{
\begin{minted}{python}
class ModelBase(object):
    @abstractmethod
    def prompt(self, **kwargs):
        ''' Call to model API '''
        pass
    @abstractmethod
    def summarize_response(self, response):
        '''Extract response from model output'''
        pass
\end{minted}
}
\vspace{-5pt}
\caption{Abstract class for implementing a new \code{Model}.}
\label{code:model}
\vspace{-5pt}
\end{listing}
% \vspace{-20pt}
\begin{listing}[t]
\footnotesize{
\begin{minted}{python}
class DatasetBase(ABC):
    @abstractmethod
    def metadata(self):
        ''' Returns metadata for the dataset. '''
        pass
    @abstractmethod
    def get_data_sample(self):
        ''' Returns a single dictionary, with 
            at least the following keys:
            'input': <input-instance>
            'label': <label> '''
        pass
    @abstractmethod
    def load_data(self, data_path):
        ''' Returns a list of dictionaries, with
            at least the following keys:
            'input': <input-instance>
            'label': <label> '''
        pass
\end{minted}
}
\vspace{-5pt}
\caption{Abstract class for implementing a new \code{Dataset}.}
\label{code:dataset}
\vspace{-5pt}
\end{listing}

\begin{listing}[h]
\footnotesize{
\begin{minted}{python}
class TaskBase(ABC):
    @abstractmethod
    def evaluate(self, true_labels, 
                predicted_labels):
        pass
\end{minted}
}
\vspace{-5pt}
\caption{%Abstract class to implement for a new Evaluation.
Abstract class for implementing a new \code{Evaluation}.}
\label{code:eval}
\vspace{-8pt}
\end{listing}
\begin{listing}[t]
\footnotesize{
\begin{minted}{python}
def config():
    return {
       'dataset':ExampleDataset,'dataset_args':{},
       'task':ExampleTask,'task_args':{},
       'model':OpenAIModel,'model_args':{},
       'general_args':{}}
def prompt(input_sample):
    ''' Construct and return the prompt following
        the model's corrsponding template'''
def post_process(response):
    ''' Apply custom post-processing on response
        and return extracted model predicition'''
\end{minted}
}
\vspace{-5pt}
\caption{Methods to implement in the \code{Asset} module.}
\label{code:asset}
\vspace{-5pt}
\end{listing}

\subsection{\code{Model Provider} module}
\label{sec:framework:models}
A \code{Model Provider} abstracts away all model-specific communication and aims to set the defaults for maximum reproducibility (for instance assign temperature value to zero by default). 
% having the temperature value be zero by default
The framework currently supports OpenAI's API, the HF Inference API, as well as FastChat and Petals for local deployments. Defining a new LLM model is straightforward by extending the \code{ModelBase} class. The initial process entails configuring essential parameters for the model setup, including factors like temperature, top\_p, etc. Furthermore, it requires the implementation of two abstract methods (shown in Listing~\ref{code:model}):
%It also requires implementing two abstract methods: 
\code{prompt} -- manages the invocation of the model API based on the input prompt; 
%is responsible for calling the model API given the input prompt, 
and \code{summarize\_response} -- extracts a summary of the response from raw model output.

\subsection{\code{Dataset} module}
\label{sec:framework:datasets}
Similar to a \code{Model provider}, a \code{Dataset} implementation aims to abstract dataset-specific code, such as loading, pre-processing, and formatting of samples. The framework comes with four generic data loaders, including Hugging Face, CSV, and JSON datasets. A custom dataset can be easily implemented by extending the \code{DatasetBase} class. When defining a new dataset, the user is required to implement at least three methods (depicted in Listing~\ref{code:dataset}). The first, \code{metadata}, %as the name implies, 
is designed to provide comprehensive metadata such as a citation or reference to the source of the dataset, its download link, and the languages it covers. The second function to be implemented in this module is \code{load\_data}, which should define a data loader capable of returning a list $S$ comprising the samples from the dataset, given the user-specified dataset path. Lastly, \code{get\_data\_sample} should be defined to return a Python dictionary representing a single sample $S_i$ extracted from the dataset.

\subsection{\code{Evaluation} module}
\label{sec:framework:tasks}
The \code{Evaluation} module aims to compute metrics and consolidate the results for a task. The framework comes with built-in support for popular task types such as Classification and Regression and is easily extendible to any custom metric by inheriting from the \code{TaskBase} class. A custom implementation can define an \code{evaluate} function for a task with specific evaluation code and metrics (see Listing~\ref{code:eval}). The function is passed two lists: the predicted labels, and true or gold labels. Its primary objective is to yield a user-defined Python dictionary comprising key-value pairs representing the outcomes of the evaluation (e.g., \code{\{``Accuracy'': accuracy value\}}).

\subsection{\code{Benchmarking Asset} module}
\label{sec:framework:assets}

The \code{Asset} module represents a benchmarking experiment, %given 
utilizing all the modules %-- datasets, models, and evaluation, 
 defined in \benchpack{} as can be seen in code snippet Listing~\ref{code:asset}. Within this module, the user should %is expected to 
 provide full configuration for the experiment, %including 
which includes %indicating 
specifying the \code{Dataset}, \code{Model}, and \code{Evaluation} modules. %to use. 
The module also enables passing model and dataset parameters. 

%This 
The \code{Asset} module must also implement the \code{prompt} function, %that 
which constructs the actual prompt to pass to the \code{Model}, %given an 
based on the input sample. %In the case of 
For scenarios involving few-shot learning, the \code{Asset} module is provided with $k$ examples to use in prompt construction. 
% selected 
%These examples are chosen by the framework from a training dataset specifically designed for the task at hand, where $k$ is a parameter controlled by the user. The training dataset is also expected to be passed by the user. 
These examples are chosen by the framework from a training dataset specified by the user, where $k$ is a parameter controlled by the user.

Finally, the \code{post\_process} function is required to be implemented to post-process
%can implement post-processing of the 
response from the model. This step is crucial because the output produced by the \code{Model} is tailored to the particular model and the prompt used, leading to potential variations across benchmarking experiments.
%This method is necessary as the response returned by the \code{Model} is specific to the model and prompt, thus can differ across different benchmarking experiments.  

\subsection{Interaction}
\label{sec:framework:running}
% \todo[inline]{"limit n" by R3}
%With 
Once the aforementioned modules are implemented, running a benchmarking experiment becomes a straightforward task, accomplished through a single command that provides access to various adjustable parameters.
%can be done with a single command exposing several parameters to set, which allows for easily configurable experiments. 
% The setup allows for effortless configuration of experiments.
The package automatically identifies the \code{Asset} to run based on wildcard search using the provided asset name. Additionally, the package determines whether to activate the few-shot setup when the number of shots is specified as a parameter (\code{-\hspace{0.1mm}-\hspace{0.1mm}n\_shots\hspace{1mm}<$k$>}). 
%provided, and detects whether to run the few-shot setup in case of passing the number of shots as a parameter (\code{---n\_shots <$k$>}). 
Furthermore, the package supports swift testing of the benchmarking asset by executing it on a small number of $n$ (\code{-\hspace{0.1mm}-\hspace{0.1mm}limit\hspace{1mm}<$n$>}) samples, which limits the run to the first $n$ samples from the dataset. An example of the command is provided below.
%It also supports quick testing of the benchamrking asset by running it over a limited number of $n$ samples from the dataset. An example command is shown next. 

\begin{lstlisting}[language=bash]
  $ python -m llmebench --filter '*AssetName*' --n_shots k --limit n --ignore_cache <benchmark-dir> <results-dir>
\end{lstlisting}

\section{Features}

\benchpack{} features a generic framework that serves a broad range of tasks and models in different learning settings (zero-/few-shot) to evaluate model performance. It enables scalable and rigorous evaluation across diverse tasks and languages while offering simplicity of implementation and flexibility in customization. 

\subsection{Modularity}
% As can be seen in Figure \ref{fig:pipeline}, 
The \benchpack{} framework, as shown in Figure \ref{fig:pipeline}, follows loosely-coupled design principles, effectively separating the data loader, models, and evaluation components. These components interact through a \code{Benchmark Driver}, ensuring a modular and flexible architecture. 

\subsection{Generality}
%\maram{Should we also mention here the granularity of data we tested: word-level, single sentence, pairs of sentences, QA pairs, etc}

% We carefully designed the framework to offer generality, offering effortless customization of tasks, data, and models.
The framework is designed to offer generality,
with effortless customization of tasks, data, and models. The framework comes with several generic data loaders such as Hugging Face datasets. Additionally, since 
users have the ability to create their own data loaders, the framework can support any standard data format. At the time of writing this paper, we have conducted tests with different formats including \textit{TSV}, \textit{CSV}, \textit{JSON}, and \textit{JSONL}.

\begin{figure*}[t]
\centering
\includegraphics[width=0.95\textwidth]{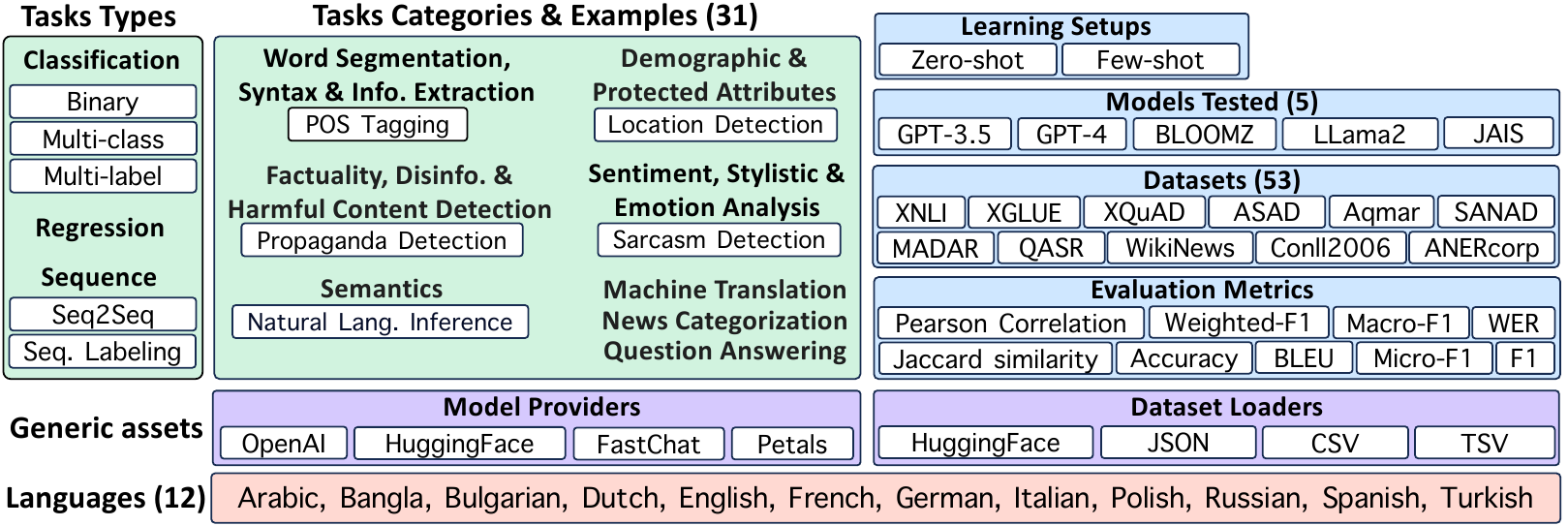}
% \vspace{-3pt}
\caption{Summary and examples of the 53 datasets, 31 tasks, 4 model providers, 5 tested models and metrics currently implemented and validated in \benchpack.}
\label{fig:tasks_datasets}
% \vspace{-5pt}
\end{figure*}

In terms of tasks, the framework demonstrates the capability to handle a diverse array of token and sequence classification tasks. Figure \ref{fig:tasks_datasets} displays the implemented task types, with built-in support for all standard generic tasks. Additionally, any new custom task can be seamlessly incorporated.

The framework also accommodates various types of models, encompassing both open and closed models, each with its own API-based options. For closed models, users can acquire API keys and endpoints from hosting providers. Meanwhile, open models can be hosted on in-house infrastructure or any accessible hosting service through APIs. We evaluated such an integration using BLOOMZ 176B 8bit version and Jais-13b (32 bit)~\cite{sengupta2023jais}, hosting it within our in-house infrastructure using Petals \cite{borzunov2022petals} and FastChat. Overall, the framework offers a high level of generality and can be readily applied and adapted to a wide range of use cases and research scenarios.

% Overall, the high level of generality that our framework can be readily applied and adapted to a wide range of use cases and research scenarios.

\subsection{Prompts}

\benchpack{} is designed to support both zero- and few-shot learning setups. The instructions in prompts can be written in any language of interest.

\paragraph{Zero-shot prompts} provide natural language instructions describing the task and expected output. 

\paragraph{Few-shot prompts} embed a small number of examples to the natural language instructions for the particular task. 
% Our framework's implementation employs user-defined, task-specific training set to select few-shot examples. 
The framework utilizes user-defined, task-specific training set to automatically select few-shot examples. 
Various strategies exist for examples selection. Among these, we have implemented maximal marginal relevance-based (MMR) \cite{carbonell1998use} selection, which has demonstrated success in previous work by \citet{ye2022complementary}. The approach computes the similarity between a test example and the example pool (e.g., training dataset) and selects $k$ examples (shots) that are both relevant and diverse. We apply the MMR technique on top of embeddings obtained from the multilingual sentence-transformers~\cite{reimers2019sentence}. However, users also have the flexibility to utilize any custom embedding model.

We have designed a highly efficient process to extract embeddings and compute few-shot examples for each test sample. Specifically, it pre-selects few-shot examples for each test sample during the initial loading stage. This design effectively eliminates the need to apply and compute the MMR score when making API calls, thus enhancing the system's overall efficiency.

The \benchpack{} framework includes $\sim$$300$ designed prompts for zero-shot and few-shot setups that have been %tested and 
validated across a variety of NLP tasks and datasets %(for more details on the \benchpack{} framework evaluation, refer to Section~\ref{sec:evaluation}). This collection serves as an excellent starting point for the community and is poised to expedite research in the field of prompt engineering.
% (for further details, refer to 
(see Section~\ref{sec:evaluation}). This collection serves as a strong starting point for the community and expedite prompt engineering research.
% is poised to expedite research in the field of prompt engineering.

\subsection{Caching}

One of the significant challenges when accessing APIs is managing timeout issues. The necessity to rerun experiments involving API calls not only requires additional effort but also increases costs. To address this problem, we have developed a caching mechanism, allowing users to bypass making API calls for samples that have already been successfully processed. Specifically, we save all intermediate outputs when processing a data sample, including the generated prompt, the raw model response and the post-processed output. On a re-run, samples that have model responses in the cache do not actually access the API, but rather load the cached response. Furthermore, this caching mechanism plays a vital role in enhancing the post-processing of the models' output, as it can be performed repeatedly without having to call the API. This feature is important, given that different models yield various types of outputs, requiring improved post-processing to align the output accurately with the reference label.
To further counter expected time-out and rate limitation issues with APIs, the framework also applies a user-configurable \textit{wait-and-retry} mechanism. This mechanism retries API calls in case of failure, maximizing the chance of receiving a successful response.

%By doing so, we improve the consistency and efficiency of our API interactions, thereby reducing both costs and effort.

\subsection{Dataset Auto-Download}
% \fd{
The framework comes with support for automatic downloading and caching of publicly available datasets, taking care of extracting and linking them correctly for any \code{Asset} that requires it. This allows a new user to quickly begin experimenting with and evaluating an existing dataset without the need to manually acquire it first.
% }

\subsection{Task Diversity}
% The framework supports and currently includes a diverse set of tasks. Covering a broad spectrum, from word-level tasks to those involving single sentences, sentence pairs, question-answer pairs, and more.  
% The range of tasks covers various NLP research tracks, as can be seen in Figure \ref{fig:tasks_datasets}.

The framework currently supports and includes a diverse set of tasks, covering a broad spectrum that ranges from word-level tasks to those involving single sentences, sentence pairs, question-answer pairs, and more. The range of tasks covers various NLP research tracks, as can be seen in Figure \ref{fig:tasks_datasets}.

% These are encompassing different NLP research tracks, from segmentation, to token classification to sentence and document classification, as can be seen in Figure \ref{fig:tasks_datasets}. 
% (both standard like POS tagging and more recent ones) 

\subsection{Language-agnostic Framework}
The \benchpack{} framework is language-agnostic. As of now, tasks for 12 languages have been incorporated, a number that will continuously grow as a result of our ongoing efforts and, ideally, with the support of the community.

\subsection{Open-source Framework}
We made %intend to make 
 \textit{\benchpack{}} accessible to the community by releasing it as open-source. This will also enable the continued growth and development of the framework within the community.

% \begin{itemize}
% \item zero and few shots
% \item Number of evaluation metrics, tasks, datasets and models already supported (and how these tasks are the most wide-spread NLP tasks)
% \item Modularity and independence between the three modules
% \item Extension ability to add new datasets, tasks, (LLM/Prompting?) models
% \item Open for contribution (open-source the package)
% \item Support and issues handling
% \item Language-agnostic
% \item Ability to add different types of datasets (single file, parallel files, etc.)
% \item Leaderboard
% \item Estimated time to run an experiment starting from adding a new benchmarking dataset and task till results (It might be beneficial to note that a junior programmer/researcher can perform this process in X minutes)
% \end{itemize}

% \section{Use Cases}
% \label{sec:use_cases}
% Use case?: our Arabic experiments, scale of team testing and using the package during our LLM paper, number of tasks and Datasets.

\subsection{Deploying Local Model}
% \todo[inline]{Can we discuss @Sabri? comment by R3}
For deploying local models, we interface \benchpack{} with FastChat framework~\cite{zheng2023judging}. The latter is an open-source project~\footnote{\url{https://github.com/lm-sys/FastChat}} for serving LLMs with a fast-growing user community. Custom chat templates of popular and new LLMs are rapidly added to the framework. This allows a proper use of instruct-tuned LLMs in inference mode, unlike other popular benchmarking frameworks such Eval-Harness~\cite{eval-harness}, which do not use chat templates. This can lead to a mismatch in the expected inputs of instruction tuned models, and put at disadvantage several LLMs, especially the ones with small sizes. Our approach for local deployment solves this issue by relying on FastChat. Huggingface models can be easily deployed in three steps after installing FastChat python package: 1) Running a model controller which plays a role of interfacing between API and model calls. 2) Running a model worker for each loaded model which manages the model in GPU and executes the prompts and returns to responses to the model worker. It is possible to load model worker with vLLM~\footnote{\url{https://github.com/vllm-project/vllm/}} enabling prompt batching for an efficient and fast inference~\cite{kwon2023efficient}. 3) Running an API server which provides a compatible interface with to OpenAI API. The local address and port of the API are set in \benchpack{} via \texttt{FASTCHAT\_*} environment variables.

\section{Evaluation of \benchpack{}}
\label{sec:evaluation}
%The \benchpack{} 
The framework has already been used across a variety of Arabic NLP tasks and datasets \cite{abdelali-2024-larabench}. This involved extensive experimentation using zero- and few-shot learning with state-of-the-art large language models, including GPT-3.5-Turbo, GPT-4, and the 8-bit version of the BLOOMZ 176B model. In Figure \ref{fig:tasks_datasets}, we provide a summary of the tasks, datasets, and models that have been implemented and evaluated. Given that our assessment of the framework was based on the current state-of-the-art NLP tasks and datasets, we implemented task- and dataset-specific metrics reported in the literature. Overall, %we 
it has been used to evaluate 31 NLP tasks, which were categorized based on ACL tracks, 53 datasets, and different model providers within 2 learning setups. All the task recipes are available within the framework.

% \section{Recipe Flow of \benchpack{}}
% \label{sec:receipe}
% \todo[inline]{Maybe go over data, task , model modules for 1 or 2 example task? like NER and QA -- same as we will use for video. See https://arxiv.org/pdf/1804.00015.pdf for Sec4.1 for example, also it would be great if we show example of prompt etc.}

\section{Related Work}
\label{sec:related_work}
%Maram: merged with intro to save space and better motivation
Efforts to assess the performance of LLMs on standard NLP tasks have been underway since the launch of ChatGPT. Notable studies, such as those by \citet{bubeck2023sparks}, \citet{bang2023multitask}, \citet{ahuja2023mega}, and \citet{hendy2023good}, have conducted large-scale experiments considering multilinguality, multimodality, low-resource languages, and a wide range of datasets and tasks.

Such large-scale evaluations require off-the-shelf and easy-to-use solutions to measure the performances of LLMs for a variety of NLP-related tasks. To evaluate OpenAI's models, the company developed the EVALs\footnote{\url{https://github.com/openai/evals}} package, which requires the dataset to be prepared in JSON format using a predefined template.
% \textcolor{red}{Currently, it does not support custom evaluation scripts.}\maram{Not sure what this means, their git says: 'Please note that we are currently not accepting Evals with custom code! While we ask you to not submit such evals at the moment, you can still submit modelgraded evals with custom modelgraded YAML files.' I also think one can clone their repo and extend for their need. We need a stronger comparison point, maybe simplicity of our package? or format-free data loaders}%, however, user can define their evluation metric for their task. 
\citet{asai2023buffet} developed an evaluation framework as a part of their cross-lingual benchmarking effort. This comprehensive framework includes 15 distinct tasks set in a few-shot learning environment across 54 languages. However, this evaluation framework was not publicly available at the time of writing this paper. OpenICL~\cite{wu-etal-2023-openicl} is another framework designed specifically for zero-shot or few-shot learning setups. It incorporates various strategies, such as random selection, heuristic methods (including BM25~\cite{robertson2009probabilistic}, TopK~\cite{liu2022makes}, and VoteK~\cite{hongjin2022selective}), and a model-based approach, to select few-shot examples. The OpenICL framework is implemented under the assumption that users will utilize HF datasets to load and evaluate models. A prompt is an important part that serves as a bridge between humans and LLMs.
% and it is known that LLMs are highly sensitive to prompts. 
To explore the research in this direction \citet{zhu2023promptbench} developed the PromptBench framework for prompt engineering.
% with a focus on the robustness of prompting.
% I removed this \footnote{\url{https://www.eleuther.ai/}} as the github url following it is broken across pages because of lack of space and this caused latex to make fig2 part of the reference (so when u click on the fig, it takes u to github) 
Eleuther-AI developed LM evaluation Harness~\cite{eval-harness}, which includes the implementation of 200 tasks and supports multiple models from the HF hub. 
% \textcolor{red}{Similar to OpenICL, this framework also assumes that the dataset will be hosted on Hugging Face.}\maram{I checked, and this package allows local datasets: please see line 516 in base.py}\maram{Maybe we can say it doesn't support local datasets out of the box}
% PhaseLLM\footnote{\url{https://github.com/wgryc/phasellm}} is another evaluation framework for LLMs that allows for the assessment of various commercial and closed models.
% , while our framework is not limited to such LLMs. 

\begin{table}[h]
\centering
\resizebox {\columnwidth}{!}{%
\begin{tabular}{@{}llllll@{}}
\toprule
\multicolumn{1}{c}{\textbf{}} & \multicolumn{3}{c}{\textbf{Customization}} & \multicolumn{2}{c}{\textbf{ICL (shot)}} \\ \midrule
\multicolumn{1}{c}{\textbf{Eval Package}} & \multicolumn{1}{c}{\textbf{Dataset}} & \multicolumn{1}{c}{\textbf{Task}} & \multicolumn{1}{c}{\textbf{Models}} & \multicolumn{1}{c}{\textbf{Zero}} & \multicolumn{1}{c}{\textbf{Few}} \\ \midrule
OpenAI evals\footnote{\url{https://github.com/openai/evals}} & Fixed & \checkmark & Any & \checkmark & \checkmark \\ 
LM Harness~\cite{eval-harness} & HF & $\checkmark$ & HF & $\checkmark$ & $\checkmark$ \\
OpenICL~\cite{wu-etal-2023-openicl} & HF & $\checkmark$ & HF, OpenAI & $\checkmark$ & $\checkmark$ \\ \midrule
\benchpack{} (Ours) & Custom & $\checkmark$ & Any & $\checkmark$ & $\checkmark$ \\ \bottomrule
\end{tabular}%
}
\vspace{-5pt}
\caption{LLMs evaluation frameworks. HF: Hugging Face
% \todo[inline]{Tab1: Add capabilities, open source?}
}
\label{tab:comparison_tabl}
\vspace{-10pt}
\end{table}

Compared to the aforementioned frameworks (summarized in Table~\ref{tab:comparison_tabl}), the \benchpack{} framework offers customization through custom dataset loaders, tasks, and models. It also supports both zero- and few-shot prompting.
%offers any sentence transformer based model selection from Hugging Face to use use MMR based example selection.
% The caching mechanism of the \benchpack{} framework is rarely available in counterparts although essential to help in saving time, cost, and enabling cost-free improvement of post-processing of the models' output. Our package also comes equipped with implementations for a wide variety of datasets, LLM models and tasks.
The caching mechanism provided by the \benchpack{} framework is a rarity among its counterparts, yet it is crucial for time and cost savings, as well as facilitating efficient post-processing enhancements to model outputs without additional expenses. 

\section{Conclusions and Future Work}
\label{sec:conclusion}

In this paper, we introduce \benchpack{}, an open-source  framework designed to facilitate the LLM benchmarking process. \benchpack{} accelerates evaluation of LLMs using pre-implemented generic datasets, tasks and model providers. In addition, it features a modular design that empowers users to integrate \textit{(i)} new tasks, \textit{(ii)} datasets, and \textit{(iii)} APIs for models. The framework incorporates caching mechanisms that effectively reduce time, costs, and effort associated with task evaluations. Currently, it includes predefined recipes covering 31 standard NLP tasks such as classification, translation, question-answering, and semantic parsing. These recipes can be readily extended to encompass novel NLP tasks, datasets, and LLM models.
% Our commitment involves continuous updates to the framework, introducing new tasks and models, thus establishing it as a valuable resource for researchers and industry practitioners engaged in LLMs evaluation and benchmarking. We also extend an invitation to the research community, encouraging active participation and contribution to this collaborative endeavor.

In future, we aim to further enhance the framework by integrating a broader array of tasks and languages. By embracing an open-source approach and encouraging active community participation, we anticipate its sustained growth through the incorporation of diverse tasks, enriched datasets, and innovative models.
Additional enhancements under consideration encompass integrating cross-validation datasets, and incorporating models featuring varied configurations (such as distinct iterations of BLOOM models). Furthermore, we are actively developing more methods for few-shot selections. Currently, our framework assumes seamless model access via APIs. 
We are committed to enhancing accessibility by enabling users to effortlessly load and utilize both offline and online models for inference purposes.
% Since we are making it open-source and allowing the community to contribute to the framework, we hope it will continue to grow with the addition of more tasks, datasets, and models. 

% Other potential improvements include the inclusion of cross-validation datasets, support for tasks with multiple datasets, and models with various configurations (e.g., different versions of BLOOM models). Finally, we plan to add more flexible ways for few-shots selection.
% In its current iteration, 

% In the current implementation, our framework assumes that models can be accessed through APIs. We will enhance it further to allow users to load and use %\textcolor{red}{
% offline and online models
% % } 
% for inference. 

\section*{Limitations}
The LLMeBench is currently limited to API calls, whether they are local or remotely hosted. It also operates under the assumption that the entire dataset can fit into memory, which may not be feasible for very large collections. Implementing iterable loading could be a viable solution to this issue, and is a feature that might be considered for future development. Additionally, many datasets come with cross-validation splits, a functionality that the framework does not currently support.

% \newpage
% \clearpage

% \section*{Limitations}
% Cross-validation seamless support

% \section*{Ethics Statement}
% We used publicly available and in-house developed datasets in our study. Any biases are unintended.
% Scientific work published at EMNLP 2023 must comply with the \href{https://www.aclweb.org/portal/content/acl-code-ethics}{ACL Ethics Policy}. We encourage all authors to include an explicit ethics statement on the broader impact of the work, or other ethical considerations after the conclusion but before the references. The ethics statement will not count toward the page limit (8 pages for long, 4 pages for short papers).

% \section*{Acknowledgments}

% \paragraph{Potential Risk}
% We do not oversee any potential risk that can result from our study. 
%Our current results are also limited to only zero-shot learning, for which performance highly depends on the prompt design and it requires significant prompt engineering effort. 

\section*{Ethics Statement}
\label{sec:ethics_statement}

Our framework incorporates publicly available datasets and relies on external models. These models may produce non-factual or potentially harmful content. Therefore, we encourage users to be aware of their interaction with the models.

\section*{Acknowledgments}
The contributions of M. Hasanain were funded by the NPRP grant 14C-0916-210015, which is provided by the Qatar National Research Fund (a member of Qatar Foundation).

% Entries for the entire Anthology, followed by custom entries
\bibliography{bib/anthology,bib/bibliography}

\begin{thebibliography}{29}
\expandafter\ifx\csname natexlab\endcsname\relax\def\natexlab#1{#1}\fi

\bibitem[{Abdelali et~al.(2024)Abdelali, Mubarak, Chowdhury, Hasanain, Mousi, Boughorbel, Abdaljalil, Kheir, Izham, Dalvi, Hawasly, Nazar, Elshahawy, Ali, Durrani, Milic-Frayling, and Alam}]{abdelali-2024-larabench}
Ahmed Abdelali, Hamdy Mubarak, Shammur~Absar Chowdhury, Maram Hasanain, Basel Mousi, Sabri Boughorbel, Samir Abdaljalil, Yassine~El Kheir, Daniel Izham, Fahim Dalvi, Majd Hawasly, Nizi Nazar, Yousseif Elshahawy, Ahmed Ali, Nadir Durrani, Natasa Milic-Frayling, and Firoj Alam. 2024.
\newblock {{LAraBench}: Benchmarking Arabic AI with Large Language Models}.
\newblock In \emph{Proceedings of the 18th Conference of the European Chapter of the Association for Computational Linguistics: Volume 1, Long Papers}, Malta. Association for Computational Linguistics.

\bibitem[{Ahuja et~al.(2023)Ahuja, Diddee, Hada, Ochieng, Ramesh, Jain, Nambi, Ganu, Segal, Ahmed, Bali, and Sitaram}]{ahuja2023mega}
Kabir Ahuja, Harshita Diddee, Rishav Hada, Millicent Ochieng, Krithika Ramesh, Prachi Jain, Akshay Nambi, Tanuja Ganu, Sameer Segal, Mohamed Ahmed, Kalika Bali, and Sunayana Sitaram. 2023.
\newblock \href {https://doi.org/10.18653/v1/2023.emnlp-main.258} {{MEGA}: Multilingual evaluation of generative {AI}}.
\newblock In \emph{Proceedings of the 2023 Conference on Empirical Methods in Natural Language Processing}, pages 4232--4267, Singapore. Association for Computational Linguistics.

\bibitem[{Asai et~al.(2023)Asai, Kudugunta, Yu, Blevins, Gonen, Reid, Tsvetkov, Ruder, and Hajishirzi}]{asai2023buffet}
Akari Asai, Sneha Kudugunta, Xinyan~Velocity Yu, Terra Blevins, Hila Gonen, Machel Reid, Yulia Tsvetkov, Sebastian Ruder, and Hannaneh Hajishirzi. 2023.
\newblock {BUFFET}: Benchmarking large language models for few-shot cross-lingual transfer.
\newblock \emph{arXiv preprint arXiv:2305.14857}.

\bibitem[{Bang et~al.(2023)Bang, Cahyawijaya, Lee, Dai, Su, Wilie, Lovenia, Ji, Yu, Chung, V.~Do, Xu, and Fung}]{bang2023multitask}
Yejin Bang, Samuel Cahyawijaya, Nayeon Lee, Wenliang Dai, Dan Su, Bryan Wilie, Holy Lovenia, Ziwei Ji, Tiezheng Yu, Willy Chung, Quyet V.~Do, Yan Xu, and Pascale Fung. 2023.
\newblock A multitask, multilingual, multimodal evaluation of {C}hat{GPT} on reasoning, hallucination, and interactivity.
\newblock In \emph{Proceedings of the 13th International Joint Conference on Natural Language Processing and the 3rd Conference of the Asia-Pacific Chapter of the Association for Computational Linguistics (Volume 1: Long Papers)}, pages 675–--718, Indonesia. Association for Computational Linguistics.

\bibitem[{Borzunov et~al.(2023)Borzunov, Baranchuk, Dettmers, Riabinin, Belkada, Chumachenko, Samygin, and Raffel}]{borzunov2022petals}
Alexander Borzunov, Dmitry Baranchuk, Tim Dettmers, Maksim Riabinin, Younes Belkada, Artem Chumachenko, Pavel Samygin, and Colin Raffel. 2023.
\newblock \href {https://doi.org/10.18653/v1/2023.acl-demo.54} {Petals: Collaborative inference and fine-tuning of large models}.
\newblock In \emph{Proceedings of the 61st Annual Meeting of the Association for Computational Linguistics (Volume 3: System Demonstrations)}, pages 558--568, Toronto, Canada. Association for Computational Linguistics.

\bibitem[{Bubeck et~al.(2023)Bubeck, Chandrasekaran, Eldan, Gehrke, Horvitz, Kamar, Lee, Lee, Li, Lundberg, Nori, Palangi, Ribeiro, and Zhang}]{bubeck2023sparks}
Sébastien Bubeck, Varun Chandrasekaran, Ronen Eldan, Johannes Gehrke, Eric Horvitz, Ece Kamar, Peter Lee, Yin~Tat Lee, Yuanzhi Li, Scott Lundberg, Harsha Nori, Hamid Palangi, Marco~Tulio Ribeiro, and Yi~Zhang. 2023.
\newblock \href {http://arxiv.org/abs/2303.12712} {Sparks of artificial general intelligence: Early experiments with {GPT}-4}.
\newblock Technical report, Microsoft Research.

\bibitem[{Carbonell and Goldstein(1998)}]{carbonell1998use}
Jaime Carbonell and Jade Goldstein. 1998.
\newblock The use of mmr, diversity-based reranking for reordering documents and producing summaries.
\newblock In \emph{Proceedings of the 21st annual international ACM SIGIR conference on Research and development in information retrieval}, pages 335--336.

\bibitem[{Chang et~al.(2023)Chang, Wang, Wang, Wu, Zhu, Chen, Yang, Yi, Wang, Wang et~al.}]{chang2023survey}
Yupeng Chang, Xu~Wang, Jindong Wang, Yuan Wu, Kaijie Zhu, Hao Chen, Linyi Yang, Xiaoyuan Yi, Cunxiang Wang, Yidong Wang, et~al. 2023.
\newblock A survey on evaluation of large language models.
\newblock \emph{arXiv preprint arXiv:2307.03109}.

\bibitem[{Dong et~al.(2023)Dong, Li, Dai, Zheng, Wu, Chang, Sun, Xu, Li, and Sui}]{dong2023survey}
Qingxiu Dong, Lei Li, Damai Dai, Ce~Zheng, Zhiyong Wu, Baobao Chang, Xu~Sun, Jingjing Xu, Lei Li, and Zhifang Sui. 2023.
\newblock A survey on in-context learning.
\newblock \emph{arXiv preprint arXiv:2301.00234}.

\bibitem[{Gao et~al.(2021)Gao, Tow, Biderman, Black, DiPofi, Foster, Golding, Hsu, McDonell, Muennighoff, Phang, Reynolds, Tang, Thite, Wang, Wang, and Zou}]{eval-harness}
Leo Gao, Jonathan Tow, Stella Biderman, Sid Black, Anthony DiPofi, Charles Foster, Laurence Golding, Jeffrey Hsu, Kyle McDonell, Niklas Muennighoff, Jason Phang, Laria Reynolds, Eric Tang, Anish Thite, Ben Wang, Kevin Wang, and Andy Zou. 2021.
\newblock \href {https://doi.org/10.5281/zenodo.5371628} {A framework for few-shot language model evaluation}.
\newblock \emph{Zenodo}.

\bibitem[{Hendy et~al.(2023)Hendy, Abdelrehim, Sharaf, Raunak, Gabr, Matsushita, Kim, Afify, and Awadalla}]{hendy2023good}
Amr Hendy, Mohamed Abdelrehim, Amr Sharaf, Vikas Raunak, Mohamed Gabr, Hitokazu Matsushita, Young~Jin Kim, Mohamed Afify, and Hany~Hassan Awadalla. 2023.
\newblock How good are {GPT} models at machine translation? a comprehensive evaluation.
\newblock \emph{arXiv preprint arXiv:2302.09210}.

\bibitem[{Hongjin et~al.(2022)Hongjin, Kasai, Wu, Shi, Wang, Xin, Zhang, Ostendorf, Zettlemoyer, Smith et~al.}]{hongjin2022selective}
SU~Hongjin, Jungo Kasai, Chen~Henry Wu, Weijia Shi, Tianlu Wang, Jiayi Xin, Rui Zhang, Mari Ostendorf, Luke Zettlemoyer, Noah~A Smith, et~al. 2022.
\newblock Selective annotation makes language models better few-shot learners.
\newblock In \emph{The Eleventh International Conference on Learning Representations}.

\bibitem[{Khondaker et~al.(2023)Khondaker, Waheed, Nagoudi, and Abdul-Mageed}]{khondaker2023gptaraeval}
Md~Tawkat~Islam Khondaker, Abdul Waheed, El~Moatez~Billah Nagoudi, and Muhammad Abdul-Mageed. 2023.
\newblock \href {https://doi.org/10.18653/v1/2023.emnlp-main.16} {{GPTA}ra{E}val: A comprehensive evaluation of {C}hat{GPT} on {A}rabic {NLP}}.
\newblock In \emph{Proceedings of the 2023 Conference on Empirical Methods in Natural Language Processing}, pages 220--247, Singapore. Association for Computational Linguistics.

\bibitem[{Kwon et~al.(2023)Kwon, Li, Zhuang, Sheng, Zheng, Yu, Gonzalez, Zhang, and Stoica}]{kwon2023efficient}
Woosuk Kwon, Zhuohan Li, Siyuan Zhuang, Ying Sheng, Lianmin Zheng, Cody~Hao Yu, Joseph~E. Gonzalez, Hao Zhang, and Ion Stoica. 2023.
\newblock Efficient memory management for large language model serving with pagedattention.
\newblock In \emph{Proceedings of the ACM SIGOPS 29th Symposium on Operating Systems Principles}.

\bibitem[{Liang et~al.(2022)Liang, Bommasani, Lee, Tsipras, Soylu, Yasunaga, Zhang, Narayanan, Wu, Kumar et~al.}]{liang2022holistic}
Percy Liang, Rishi Bommasani, Tony Lee, Dimitris Tsipras, Dilara Soylu, Michihiro Yasunaga, Yian Zhang, Deepak Narayanan, Yuhuai Wu, Ananya Kumar, et~al. 2022.
\newblock Holistic evaluation of language models.
\newblock \emph{arXiv preprint arXiv:2211.09110}.

\bibitem[{Liu et~al.(2022)Liu, Shen, Zhang, Dolan, Carin, and Chen}]{liu2022makes}
Jiachang Liu, Dinghan Shen, Yizhe Zhang, William~B Dolan, Lawrence Carin, and Weizhu Chen. 2022.
\newblock What makes good in-context examples for gpt-3?
\newblock In \emph{Proceedings of Deep Learning Inside Out (DeeLIO 2022): The 3rd Workshop on Knowledge Extraction and Integration for Deep Learning Architectures}, pages 100--114.

\bibitem[{OpenAI(2023)}]{openai2023gpt4}
OpenAI. 2023.
\newblock \href {http://arxiv.org/abs/2303.08774} {{GPT}-4 technical report}.
\newblock Technical report, OpenAI.

\bibitem[{Radford et~al.(2019)Radford, Wu, Child, Luan, Amodei, Sutskever et~al.}]{radford2019language}
Alec Radford, Jeffrey Wu, Rewon Child, David Luan, Dario Amodei, Ilya Sutskever, et~al. 2019.
\newblock Language models are unsupervised multitask learners.
\newblock \emph{OpenAI blog}, 1(8):9.

\bibitem[{Reimers and Gurevych(2019)}]{reimers2019sentence}
Nils Reimers and Iryna Gurevych. 2019.
\newblock Sentence-bert: Sentence embeddings using siamese bert-networks.
\newblock In \emph{Proceedings of the 2019 Conference on Empirical Methods in Natural Language Processing and the 9th International Joint Conference on Natural Language Processing (EMNLP-IJCNLP)}, pages 3982--3992.

\bibitem[{Robertson et~al.(2009)Robertson, Zaragoza et~al.}]{robertson2009probabilistic}
Stephen Robertson, Hugo Zaragoza, et~al. 2009.
\newblock The probabilistic relevance framework: Bm25 and beyond.
\newblock \emph{Foundations and Trends{\textregistered} in Information Retrieval}, 3(4):333--389.

\bibitem[{Scao et~al.(2022)Scao, Fan, Akiki, Pavlick, Ili{\'c}, Hesslow, Castagn{\'e}, Luccioni, Yvon, Gall{\'e} et~al.}]{scao2022bloom}
Teven~Le Scao, Angela Fan, Christopher Akiki, Ellie Pavlick, Suzana Ili{\'c}, Daniel Hesslow, Roman Castagn{\'e}, Alexandra~Sasha Luccioni, Fran{\c{c}}ois Yvon, Matthias Gall{\'e}, et~al. 2022.
\newblock {BLOOM}: A 176b-parameter open-access multilingual language model.
\newblock \emph{arXiv preprint arXiv:2211.05100}.

\bibitem[{Sengupta et~al.(2023)Sengupta, Sahu, Jia, Katipomu, Li, Koto, Afzal, Kamboj, Pandit, Pal et~al.}]{sengupta2023jais}
Neha Sengupta, Sunil~Kumar Sahu, Bokang Jia, Satheesh Katipomu, Haonan Li, Fajri Koto, Osama~Mohammed Afzal, Samta Kamboj, Onkar Pandit, Rahul Pal, et~al. 2023.
\newblock Jais and jais-chat: Arabic-centric foundation and instruction-tuned open generative large language models.
\newblock \emph{arXiv preprint arXiv:2308.16149}.

\bibitem[{Srivastava et~al.(2023)Srivastava, Rastogi, Rao, Shoeb, Abid, Fisch, Brown, Santoro, Gupta, Garriga-Alonso et~al.}]{srivastava2023beyond}
Aarohi Srivastava, Abhinav Rastogi, Abhishek Rao, Abu Awal~Md Shoeb, Abubakar Abid, Adam Fisch, Adam~R Brown, Adam Santoro, Aditya Gupta, Adri{\`a} Garriga-Alonso, et~al. 2023.
\newblock Beyond the imitation game: Quantifying and extrapolating the capabilities of language models.
\newblock \emph{Transactions on Machine Learning Research}.

\bibitem[{Wei et~al.(2022)Wei, Wang, Schuurmans, Bosma, Xia, Chi, Le, Zhou et~al.}]{weichain}
Jason Wei, Xuezhi Wang, Dale Schuurmans, Maarten Bosma, Fei Xia, Ed~H Chi, Quoc~V Le, Denny Zhou, et~al. 2022.
\newblock Chain-of-thought prompting elicits reasoning in large language models.
\newblock In \emph{Advances in Neural Information Processing Systems}.

\bibitem[{Wu et~al.(2023)Wu, Wang, Ye, Wu, Feng, Xu, and Qiao}]{wu-etal-2023-openicl}
Zhenyu Wu, Yaoxiang Wang, Jiacheng Ye, Zhiyong Wu, Jiangtao Feng, Jingjing Xu, and Yu~Qiao. 2023.
\newblock \href {https://aclanthology.org/2023.acl-demo.47} {{O}pen{ICL}: An open-source framework for in-context learning}.
\newblock In \emph{Proceedings of the 61st Annual Meeting of the Association for Computational Linguistics (Volume 3: System Demonstrations)}, pages 489--498, Toronto, Canada. Association for Computational Linguistics.

\bibitem[{Ye et~al.(2023)Ye, Iyer, Celikyilmaz, Stoyanov, Durrett, and Pasunuru}]{ye2022complementary}
Xi~Ye, Srinivasan Iyer, Asli Celikyilmaz, Veselin Stoyanov, Greg Durrett, and Ramakanth Pasunuru. 2023.
\newblock \href {https://doi.org/10.18653/v1/2023.findings-acl.273} {Complementary explanations for effective in-context learning}.
\newblock In \emph{Findings of the Association for Computational Linguistics: ACL 2023}, pages 4469--4484, Toronto, Canada. Association for Computational Linguistics.

\bibitem[{Zhao et~al.(2023)Zhao, Zhou, Li, Tang, Wang, Hou, Min, Zhang, Zhang, Dong et~al.}]{zhao2023survey}
Wayne~Xin Zhao, Kun Zhou, Junyi Li, Tianyi Tang, Xiaolei Wang, Yupeng Hou, Yingqian Min, Beichen Zhang, Junjie Zhang, Zican Dong, et~al. 2023.
\newblock A survey of large language models.
\newblock \emph{arXiv preprint arXiv:2303.18223}.

\bibitem[{Zheng et~al.(2023)Zheng, Chiang, Sheng, Zhuang, Wu, Zhuang, Lin, Li, Li, Xing et~al.}]{zheng2023judging}
Lianmin Zheng, Wei-Lin Chiang, Ying Sheng, Siyuan Zhuang, Zhanghao Wu, Yonghao Zhuang, Zi~Lin, Zhuohan Li, Dacheng Li, Eric Xing, et~al. 2023.
\newblock Judging llm-as-a-judge with mt-bench and chatbot arena.
\newblock \emph{arXiv preprint arXiv:2306.05685}.

\bibitem[{Zhu et~al.(2023)Zhu, Wang, Zhou, Wang, Chen, Wang, Yang, Ye, Gong, Zhang et~al.}]{zhu2023promptbench}
Kaijie Zhu, Jindong Wang, Jiaheng Zhou, Zichen Wang, Hao Chen, Yidong Wang, Linyi Yang, Wei Ye, Neil~Zhenqiang Gong, Yue Zhang, et~al. 2023.
\newblock {PromptBench}: Towards evaluating the robustness of large language models on adversarial prompts.
\newblock \emph{arXiv preprint arXiv:2306.04528}.

\end{thebibliography}
\bibstyle{acl_natbib}

% \appendix

% \newpage
% \clearpage
% \section*{Appendix}
% \label{sec:appendix}
% \appendix
% \input{appendix}

\end{document}